\documentclass{svproc}

\usepackage{url}

\usepackage{cite}
\usepackage{amsmath}
\usepackage{orcidlink}
\usepackage{algorithm}
\usepackage{algpseudocode}
\usepackage{graphicx}
\usepackage{microtype}

\begin{document}
\mainmatter
\title{Towards Quadruped-Provided Localization and Active Tracking for Micro-UAVs}
\titlerunning{Quadruped-Provided Localization for Micro-UAVs}
\author{Alejandro Lorite Mora\inst{1,2,3}\orcidlink{0000-0002-8704-1248} \and Andrés Faíña\inst{1}\orcidlink{0000-0001-7288-6988}}
\authorrunning{Lorite Mora and Faíña}
\tocauthor{Alejandro Lorite Mora and Andrés Faíña}
\institute{IT University of Copenhagen, Rued Langgaards Vej 7, Copenhagen 2300, Denmark,\\
\email{\{lori, anfv\}@itu.dk}\\
\and Helix Lab, Campus Kalundborg 3, 4400 Kalundborg, Denmark\\
\and Novo Nordisk A/S, Hallas Alle 1, 4400 Kalundborg, Denmark}

\maketitle

\begin{abstract}
Micro unmanned aerial vehicles (micro-UAVs) are small enough to reach confined spaces that larger robots cannot access, but too small to carry the sensing and computing power required for autonomous flight. We move the localization stack entirely off the aerial platform onto a quadruped robot with a 7-degree-of-freedom (DOF) arm, which supplies the micro-UAV (27\,g bare, 42\,g with fiducial markers) its full 6-DOF pose. A camera at the arm's end-effector detects AprilTag fiducial markers on the drone and composes that observation with the quadruped's own self-localization to place the drone in a shared map frame---so the ground robot \emph{localizes} its partner, rather than only tracking it relative to the camera. The arm acts as an actively-controlled observer, repositioning to keep the drone in view as both robots move; the drone carries only an inertial measurement unit and fuses the external pose to fly commanded setpoints. In lab flights the external pose is accurate to 12--16\,mm, enough to fly the drone autonomously within 2--5\,cm of motion-capture-fed control. Having the quadruped actively follow the drone reduces the tracking error from 11.0\,cm to 6.9\,cm by holding the camera in the close range, where the markers are most accurate.
\end{abstract}

\section{INTRODUCTION}

Traversing narrow passages and highly constrained environments generally necessitates the use of small robotic platforms. However, reducing the robot's size also limits the payload capacity available for sensors, onboard computation, and communication systems. This trade-off is evident in current aerial inspection applications, which are predominantly performed using relatively large unmanned aerial vehicles (UAVs). These platforms can carry advanced sensing suites and powerful onboard processors, enabling autonomous navigation and operation in complex environments. Their large size, however, prevents them from accessing the confined or hazardous spaces that applications such as search-and-rescue missions and infrastructure inspection often require.

In contrast, micro-scale UAVs possess the agility and compact form factor required to access narrow and cluttered spaces that are inaccessible to larger platforms. Nevertheless, their limited payload capacity significantly restricts their sensing and computational capabilities, preventing fully autonomous navigation in such environments. As a result, micro-UAVs often rely on external infrastructure, offboard computation, or human assistance to safely traverse and operate within confined spaces.

A micro-UAV that cannot localize itself must obtain its pose either by adding onboard sensing or by receiving it from an external source. Scaling the payload up is the usual answer when a global navigation satellite system (GNSS) is unavailable, as in indoor drone flight~\cite{jarraya_gnss_denied_2025}: visual-inertial odometry (VIO) and light detection and ranging (LiDAR) simultaneous localization and mapping (SLAM) are the standard methods at the kilogram scale~\cite{qin_vinsmono_2018,geneva_openvins_2020,xu_fastlio2_2022}, but they degrade in featureless or poorly-lit environments and cannot run on sub-30\,g platforms, because the stereo or LiDAR sensors and the VIO/SLAM computation exceed the payload, computing power, and power a micro-UAV can carry. Micro-UAVs instead rely on lighter methods such as optical flow or learned visual navigation~\cite{palossi_nanodrone_2019}, which also degrade where visual texture is weak, as it often is in the confined spaces micro-UAVs are meant to reach.

The other option is to receive the pose from an external source. Micro-UAVs commonly rely on external infrastructure for this. Motion capture (MOCAP) with Crazyswarm2~\cite{honig_crazyswarm2} provides centimeter accuracy but requires a pre-installed array of cameras. Ultra-wideband (UWB) localization, which estimates range from fixed radio anchors, shares the same pre-setup constraint. Fiducial markers are the standard low-cost option. AprilTag~\cite{olson_apriltag_2011} is the most widely used, and its accuracy and failure modes are well characterized~\cite{kalaitzakis_fiducial_2021}. Recent open micro-UAV frameworks build on fiducial-marker localization for the same reason~\cite{chiun_star_2024}; but in all such work the observing sensor is fixed or rigidly mounted to the platform being localized. Repositioning the observer instead---moving it to keep the target in view---is an instance of active perception~\cite{bajcsy_active_2018} and, at the control level, visual servoing~\cite{chaumette_vs_2006}, but both classically serve the observer's own task rather than sustaining the external localization of a separate, non-sensing robot.

A common alternative is to let a ground vehicle help localize the UAV, providing mapping and localization support. Such cooperative UAV and unmanned ground vehicle (UGV) teams have been studied for GNSS-denied navigation with complementary sensing across platforms. UWB cooperative localization~\cite{queralta_uwb_2020} requires radios on both platforms. The most closely related work is by Akhihiero et al.~\cite{depetrillo_ugv_uav_2024}, where a UGV provides mapping, self-localization, and UAV state estimation in subterranean environments using UGV LiDAR and UWB, but the formulation requires either UAV or UGV to be stationary during state updates. Gross et al.~\cite{gross_aerial_2019} make similar architectural choices with UGV LiDAR plus stereo and IMU on the UAV. In all of these the UAV still carries its own localization sensors---stereo, inertial measurement unit (IMU), altimeter---and the ground robot only refines that estimate rather than replacing it. Despite the growing interest in cooperative robotic systems, the problem of enabling localization for sensing-constrained micro-UAVs through information provided by a ground robot remains insufficiently studied.

In our system the micro-UAV carries only an IMU, with no exteroceptive sensing and no way to localize itself, so the ground partner supplies its entire absolute 6-DOF pose while both robots move. The drone's onboard extended Kalman filter (EKF) fuses that external pose with its IMU to follow position setpoints. A quadruped with a 7-DOF arm estimates the drone's pose externally from observing AprilTags on the drone, with the arm acting as an actively-controlled perception platform that repositions to keep the drone in view. The estimate is produced in the quadruped's map frame rather than only relative to the camera, so the ground robot localizes the drone rather than merely tracking it. That pose is ready for a mapping-and-planning module to consume; integrating that module is the natural next step and is not evaluated here.

Our contributions are (i)~the integrated localization-and-tracking system---AprilTag-anchored external pose estimation with an actively-controlled arm---for a micro-UAV with no localization sensing, deployed end-to-end in the Robot Operating System (ROS\,2); and (ii)~preliminary validation of the approach in lab settings. The external estimate localizes the drone to 12--16\,mm once a constant calibration bias is removed. This is accurate enough to fly the drone autonomously along commanded trajectories, within a few centimeters of motion-capture-fed control. Moreover, active arm tracking improves rather than degrades the pose, because it holds the drone in the close range where fiducial accuracy is higher, so the quadruped's own motion helps rather than harms the estimate it provides.

\section{METHOD}

\subsection{System Overview}
\label{sec:overview}

We use a quadruped robot with an arm that provides the perception capabilities for drone localization. All sensing, estimation, and control run on the quadruped, while the drone is only a thin actuator that executes setpoints.

\begin{figure}[!htbp]
  \centering
  \includegraphics[width=0.8\columnwidth,height=0.32\textheight,keepaspectratio]{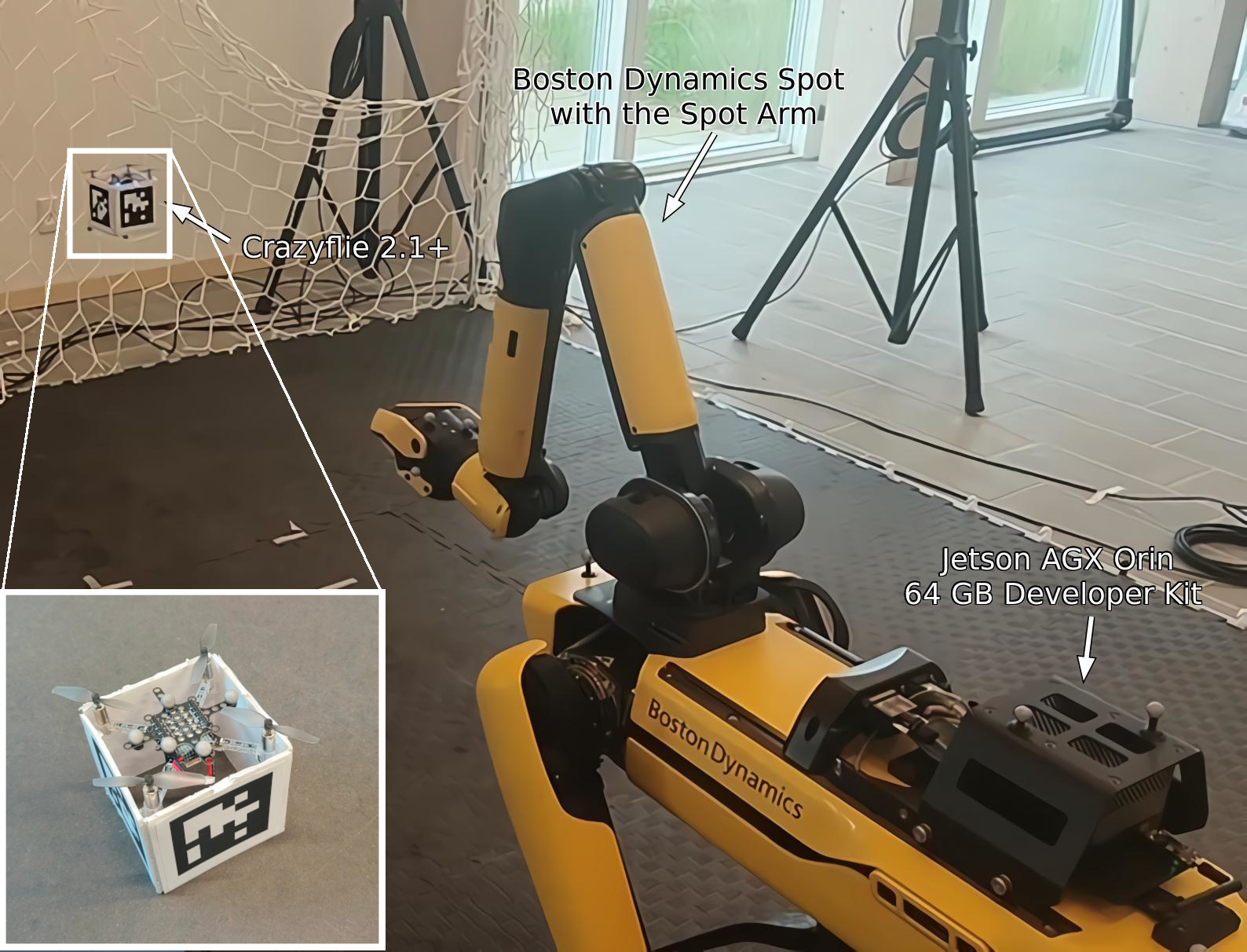}
  \caption{The integrated hardware. The Boston Dynamics Spot carries the Spot Arm (7\,DOF including the gripper)---whose end-effector camera supplies the AprilTag observations of the drone that drive the perception path---and the Jetson AGX Orin on its payload rails. The Crazyflie~2.1+ with its four-AprilTag chassis is the aerial platform the arm camera observes.}
  \label{fig:system}
\end{figure}

The ground platform is a Boston Dynamics Spot~\cite{boston_dynamics_spot} with the Spot Arm and a Jetson AGX Orin 64\,GB Developer Kit (Fig.~\ref{fig:system}). It localizes itself from its five stereo RGBD cameras, a gray-scale camera pointing to the ground, and an IMU. The aerial platform is a Crazyflie~2.1+ (27\,g bare, 92\,$\times$\,92\,$\times$\,29\,mm) with no positioning sensors; it carries a custom chassis with four 36h11 AprilTags (58.5\,$\times$\,58.5\,mm) on its lateral faces for visibility from any azimuth, bringing the flight mass to 42\,g. A Crazyradio PA (2.4\,GHz, USB) plugged into the Jetson removes the need for a separate ground station. To report position accuracy we use an OptiTrack MOCAP system of eight PrimeX 13W cameras covering a 4\,$\times$\,4\,$\times$\,2.5\,m volume; both robots carry reflective markers tracked by it.

\begin{figure}[!htbp]
  \centering
  \includegraphics[width=\textwidth]{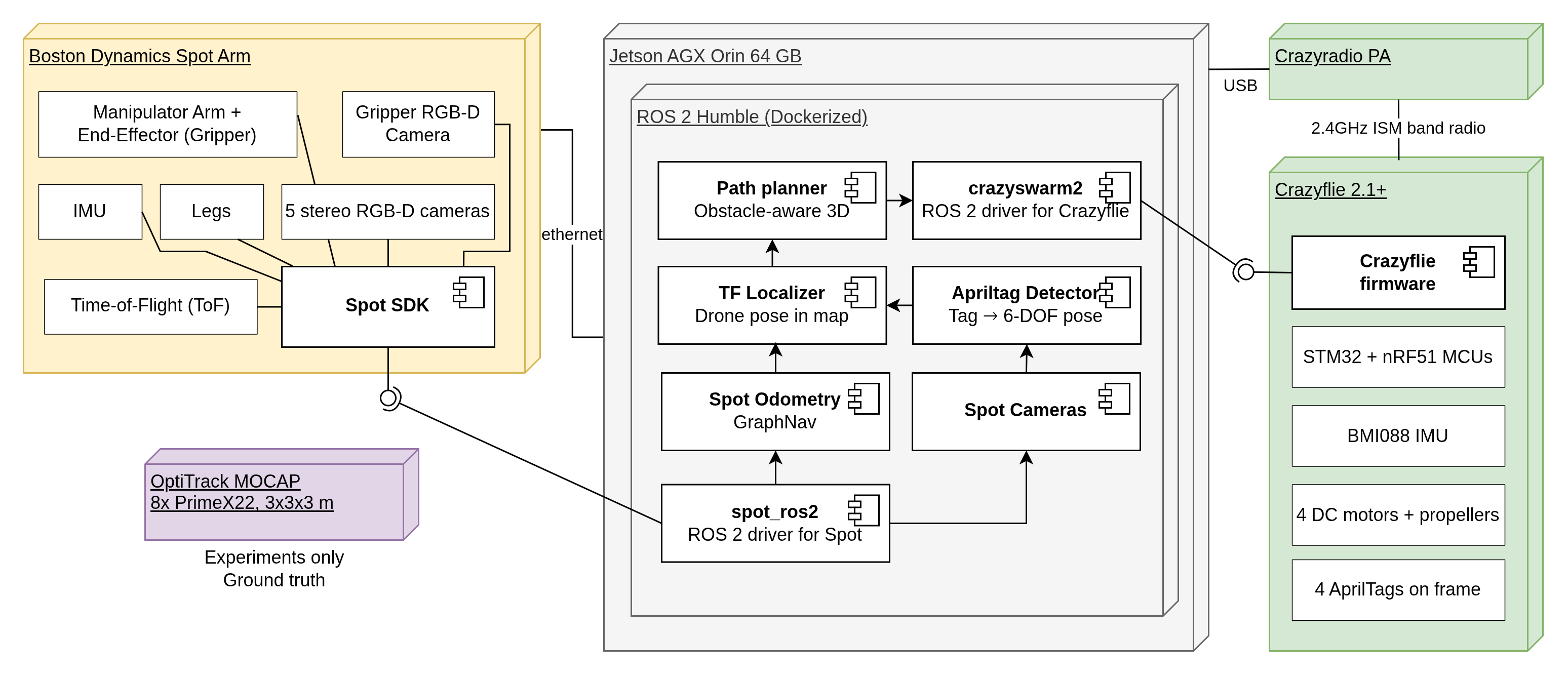}
  \caption{System architecture. A \emph{perception path} (the quadruped's sensors and the
  arm-mounted camera through the quadruped's SLAM, the \texttt{apriltag\_relocalizer}, and
  AprilTag drone-pose estimation) and a \emph{command path} (trajectory commander
  through the micro-UAV control stack and radio link to the drone) run concurrently in a
  single ROS\,2 domain; control consumes the composed $T_{\text{map}\to\text{drone}}$
  and the drone publishes no pose of its own.}
  \label{fig:arch}
\end{figure}

All software runs in ROS\,2 Humble~\cite{macenski_ros2_2022} in an Ubuntu 22.04 Docker container deployed on the Jetson PC, with packages for the quadruped's ROS\,2 driver~\cite{bdai_spot_ros2}, the micro-UAV control stack~\cite{honig_crazyswarm2}, the AprilTag detector (\texttt{apriltag\_ros}), and our integration code. Two paths run concurrently (Fig.~\ref{fig:arch}): a \emph{perception path} that fuses the quadruped's sensors and the arm-mounted camera into the drone's pose, and a \emph{command path} that carries trajectory setpoints to the drone. The control system consumes a single composed transform, $T_{\text{map}\to\text{drone}}$ (section \S\ref{sec:localization}). Because this pose is already expressed in the map frame, it is also what an on-board planner would need, so a mapping-and-planning module can be added later without changing the perception path.

\subsection{Drone Localization in the Map Frame}
\label{sec:localization}

The goal of this stage is to know where the drone is---its full 6-DOF pose, position and orientation---expressed in the same global map frame that Spot uses to navigate, so that the aerial and ground platforms share one coordinate system. Since the drone carries no positioning sensors, this pose is recovered entirely from the arm camera's view of the on-board AprilTags together with Spot's own self-localization. The estimate is assembled as a chain of coordinate transforms, each linking one frame to the next:
\begin{equation}
T_{\text{map} \to \text{drone}} = T_{\text{map} \to \text{vision}} \cdot T_{\text{vision} \to \text{body}} \cdot T_{\text{body} \to \text{cam}} \cdot T_{\text{cam} \to \text{tag}} \cdot T_{\text{tag} \to \text{drone}}.
\label{eq:tf_chain}
\end{equation}

\textbf{Fiducial detection.} Detecting the drone's AprilTags yields the camera-to-drone transform $T_{\text{cam} \to \text{drone}}$, the final two links of the chain in Eq.~\ref{eq:tf_chain}. With markers on four lateral faces the camera sees one or two tags at a time. When more than one is visible we use the single strongest detection by decision-margin followed by Perspective-n-Point (PnP)-covariance trace, then compose with a static transform to obtain $T_{\text{cam} \to \text{drone}}$. This process runs on the CPU.

\textbf{Pose stream and watchdog.} The pipeline emits one map-frame pose per detection and does not extrapolate between detections, so during a detection gap it simply publishes nothing. Tolerance to short gaps instead comes from the drone's onboard EKF, which coasts on its IMU until the next external pose arrives. A safety watchdog complements this: if the drone is airborne and no pose update arrives within a short timeout, it commands an immediate landing, so the drone is never left flying without a position estimate when it leaves the camera's view.

\subsection{Active Tracking}
\label{sec:tracking}

The quadruped's arm acts as an actively-controlled perception platform, repositioning to keep the drone in the hand camera's field of view. When the drone moves away from the center of the view, a new optimal camera position is calculated. Then, inverse kinematics (IK) is solved with the Spot Arm IK service to compute the required joint angles. The onboard locomotion controller handles foothold planning and dynamic balance. Finally, the quadruped walks and moves the arm simultaneously to re-center the drone in the image.

Let $\mathbf{p}_\text{d}$ be the predicted drone position at the next control step---estimated by fitting a velocity to the recent detections and extrapolating, which also lets the arm keep tracking briefly through a detection gap---and $\mathbf{p}_\text{c}$ the current hand-camera position. The controller maintains two conditions: the camera's optical axis points at $\mathbf{p}_\text{d}$, and the camera-to-drone distance $\|\mathbf{p}_\text{d} - \mathbf{p}_\text{c}\|$ is driven toward a fixed stand-off $d_\text{offset}\approx0.9$\,m. Larger distances remain usable---accuracy degrades only gradually with distance---so this is a preferred operating point rather than a hard limit, and it depends on the camera resolution, the AprilTag size and family, and lighting conditions. Algorithm~\ref{alg:tracking} summarizes the control loop.

\begin{algorithm}[!htbp]
\caption{Quadruped drone-tracking control loop}
\label{alg:tracking}
\begin{algorithmic}[1]
\Loop \Comment{triggered on AprilTag detection}
  \State $T^*_\text{hand} \gets$ AprilTag's frame with a fixed offset $d_\text{offset}$ along AprilTag's Z-axis
  \State $\mathbf{p}_\text{hand} \gets$ odom frame of $T^*_\text{hand}$ via TF lookup (odom $\to$ tag)
  \State $d \gets \|\mathbf{p}_\text{hand} - \mathbf{p}_\text{body}\|$ \Comment{distance to body}
  \If{$d < d_\text{min\_safe}$}
    \State skip update and continue
  \EndIf
  \If{$d > d_\text{max\_reach}$}
    \State clamp $\mathbf{p}_\text{hand}$ to $d_\text{max\_reach}$ sphere around body
  \EndIf
  \If{body\_follow enabled}
    \State command arm and body to a 6D pose $T^*_\text{hand}$ with 2\,s duration \Comment{the body moves to re-center the arm and accommodate the arm's reach}
  \Else
    \State command arm to a 6D pose $T^*_\text{hand}$ with 2\,s duration \Comment{the body stays put and the arm alone tracks the drone}
  \EndIf
  \State record detection timestamp for watchdog
\EndLoop
\State \textbf{watchdog:} if no detection for $t_\text{timeout}$ (typ. 2 s), stop servo
\end{algorithmic}
\end{algorithm}

\subsection{Experimental Setup}

All experiments run in our indoor robotics lab within a $3\times3\times2$\,m working volume. Eight OptiTrack PrimeX~13W cameras, mounted around a $4\times4\times2.5$\,m volume and aimed inward, track this space and provide 6-DOF ground truth at 120\,Hz. We use this ground truth in two ways: as the evaluation reference in every experiment, and---in selected runs (Experiment~2)---as the external pose source fed to the drone, so it can be compared head-to-head against the AprilTag estimate. Every run records the following data: fiducial detections, the composed $T_{\text{map}\to\text{drone}}$ pose, joint states, and the OptiTrack streams---together with a description of the run's configuration (pose source, AprilTag backend, hand-camera resolution, and mocap-to-odom calibration).

\section{EXPERIMENTS AND RESULTS}

The three experiments build on one another to locate the system's error budget. Experiment~1 holds the quadruped stationary and measures the intrinsic accuracy of the AprilTag pose against motion-capture ground truth, as a function of camera-to-drone distance and image resolution. Experiment~2 then flies the drone under the AprilTag estimate and asks whether that pose is accurate enough to control the flight, comparing its trajectory tracking against the same flight flown under motion capture. Experiment~3 finally lets the quadruped move---the arm, and then the body, actively following the drone---to test whether the quadruped's own motion improves or degrades the pose it supplies.

All three experiments use the same fixed drone trajectory: a horizontal sweep in front of the stationary quadruped (Fig.~\ref{fig:exp2}a). At each of three heights ($z = 0.8$, $0.55$, $0.3$\,m) the drone sweeps $\pm0.5$\,m laterally, then steps down $0.25$\,m to the next height, and then comes back to the other side. Because the drone moves through this grid while the quadruped's hand camera stays put, each pass views the drone from a different range---spanning $\approx$1.75--2.75$\,$m. This camera-to-drone distance is the $x$-axis of the accuracy plots in Experiments 1 and 2.

\subsection{Experiment 1: AprilTag Pose Estimation Accuracy}
\label{sec:exp1}

Before trusting the AprilTag pose to control a flight, we measure how accurate it is on its own. The quadruped stands still with its arm aimed at the working volume; the drone flies the previously mentioned trajectory under motion-capture control, repeated at two hand-camera resolutions (640$\times$480 and 1280$\times$720) over 29 flights (134.8k samples). At each timestep we compare the composed $T_{\text{map}\to\text{drone}}$ pose---the one downstream planning would use---against the OptiTrack drone pose.

Because we measured the camera and tag extrinsics by hand, the raw error contains a fixed setup offset that is the same in every frame. We therefore report two numbers: the \emph{raw} error (what a user gets as delivered), and the error after removing that constant offset by standard rigid alignment\footnote{Position by Horn~\cite{horn_1987}, orientation by Markley~\cite{markley_2007}, fit per flight---the alignment used by SLAM/VIO tools such as \texttt{evo\_ape}.} (\emph{aligned}), which isolates the sensor's intrinsic accuracy.

\begin{figure}[!htbp]
  \centering
  \includegraphics[width=\textwidth]{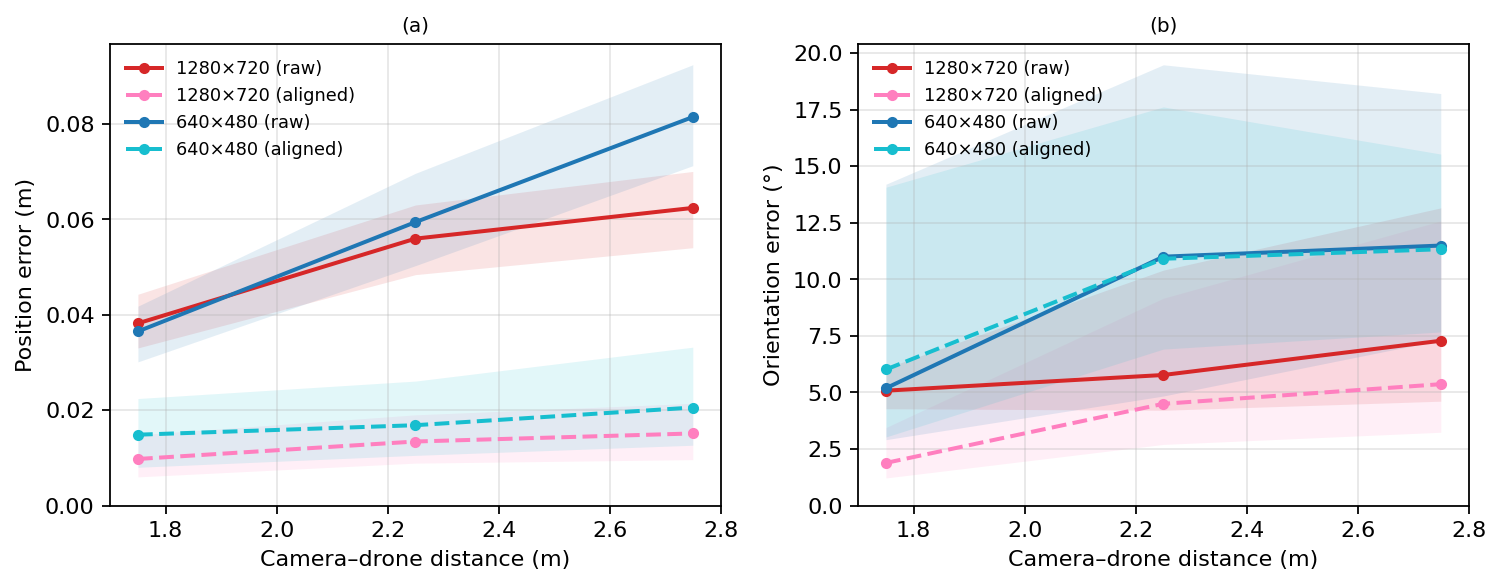}
  \caption{Experiment 1: AprilTag vs. MOCAP position and orientation error by camera resolution (raw data and Horn/Markley aligned).}
  \label{fig:exp1}
\end{figure}

The raw position error is $\approx$50\,mm and nearly identical at both resolutions, because it is dominated by the $\approx$46\,mm fixed offset. Once that offset is removed, a clear resolution effect appears: the aligned median position error is 12\,mm at 1280$\times$720 versus 16\,mm at 640$\times$480, and the high-resolution camera is $\approx$2.5$\times$ more accurate in orientation (3.4$^{\circ}$ vs 8.8$^{\circ}$); both errors grow gradually with distance (aligned position p95 30\,mm vs 45\,mm) (Fig.~\ref{fig:exp1}). The orientation gap is genuinely resolution-limited, not a calibration artifact: 640$\times$480 orientation barely changes with alignment (8.0$^{\circ}$ raw, 8.8$^{\circ}$ aligned), i.e.\ it is pixel-level tag-corner quantization the higher resolution avoids.

Per detection, 1280$\times$720 resolution is more accurate---but it cannot sustain the rate a control loop needs, for two compounding reasons. First, the Spot SDK streams the $\approx$3$\times$ larger 1280$\times$720 frames over gRPC (the Remote Procedure Call framework used by the SDK) at roughly half the rate of 640$\times$480 ($\approx$28\,Hz vs $\approx$57\,Hz at the camera driver), so the pipeline is starved of frames before any processing. Second, image rectification and AprilTag detection both scale with pixel count, dropping a further fraction of those frames. The net AprilTag pose rate is only $\approx$17\,Hz at 1280$\times$720 versus $\approx$40\,Hz at 640$\times$480, and 15.7\% of inter-detection gaps exceed 100\,ms---during which the external pose stalls and the drone's EKF coasts---versus 1.1\% respectively. For a pose-fed control loop this throughput cost outweighs the millimeter-level accuracy gain at our measured distances, so we run all closed-loop experiments at 640$\times$480 resolution.

\subsection{Experiment 2: Tracking Error -- MOCAP vs.\ AprilTag Pose Source}
\label{sec:exp2}

Experiment~1 measured the AprilTag pose in isolation; here we ask the practical question---is it accurate enough to \emph{fly} the drone? The quadruped still stands still; the drone flies the same sweep at 640$\times$480 resolution, once with its position controller fed by motion capture and once by the AprilTag estimate---three runs per pose source at each of three $x$-axis distance bins (the same grid as Experiment~1), for 18 flights (84.8k samples).

\begin{figure}[!htbp]
  \centering
  \includegraphics[width=\columnwidth]{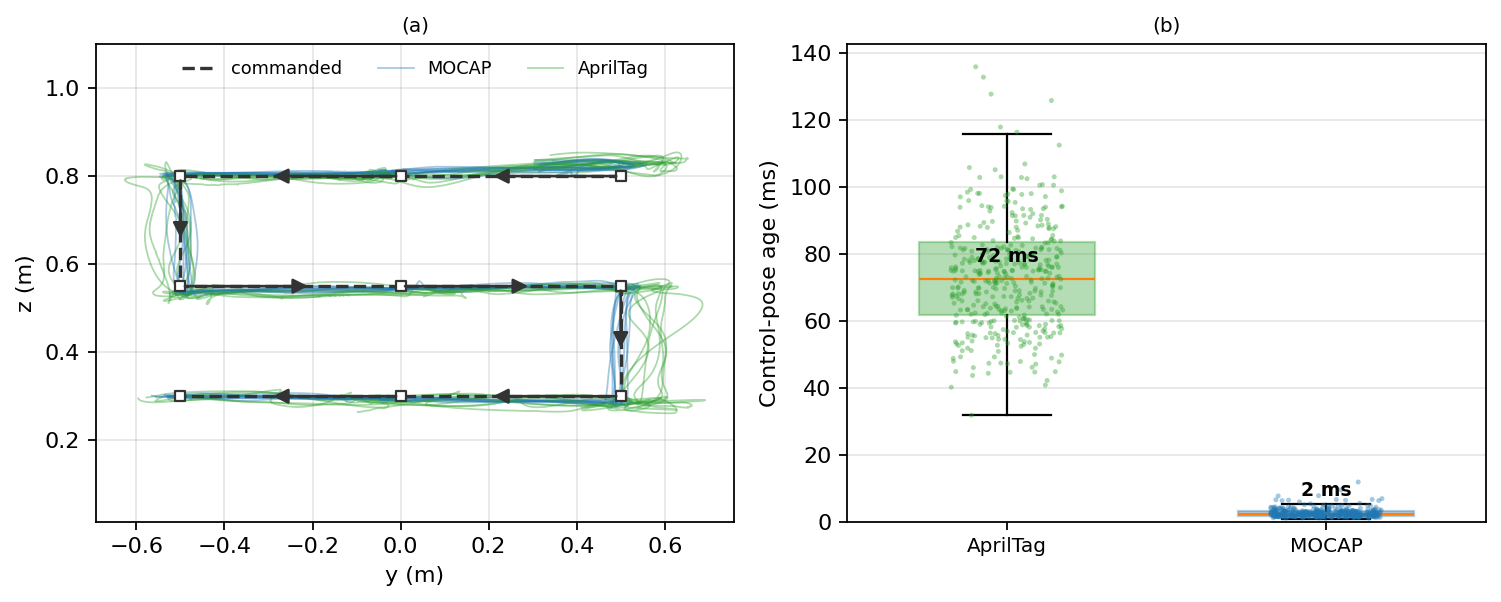}
  \caption{Experiment 2: MOCAP vs AprilTag-controlled flight. (a) Commanded trajectory (dashed) and per-waypoint median flown position with IQR; AprilTag overshoots the y = ±0.5 turn-arounds. (b) Age of the control pose reaching the drone: AprilTag is a median 72 ms stale vs 2 ms for MOCAP, explaining the overshoot.}
  \label{fig:exp2}
\end{figure}

The metric is the drone's trajectory error against its commanded waypoints (MOCAP ground truth). Unlike Experiment~1, \textbf{no extrinsic alignment is applied}: this is a closed-loop control error against a known plan, which has no sensor-calibration offset to remove, so all values are raw. Because the error is heavy-tailed---waypoint-transition transients inflate the mean and root-mean-square error (RMSE)---we report the median and interquartile range (IQR) (Fig.~\ref{fig:exp2}).

AprilTag-flown tracking error rose with distance, from a median 7.8\,cm at 1.5--2.0\,m to 11.0\,cm at 2.5--3.0\,m, against a near-constant $\approx$5.5\,cm under motion capture---a 2--5\,cm penalty that widens with distance, mirroring the accuracy trend of Experiment~1. The AprilTag estimate is therefore good enough to fly the trajectory autonomously, at a modest and predictable cost.

\subsection{Experiment 3: Active Tracking}
\label{sec:exp3}

In deployment the quadruped does not stand still---it actively aims its camera at the drone. We test whether that self-motion helps or hurts the pose it supplies. The arm tracker and base-follow controller of Algorithm~\ref{alg:tracking} run on the AprilTag pose, closing the full loop (perception, tracking, control). We compare three conditions on the same commanded drone trajectory (8 flights, 640$\times$480): \emph{still} (arm and base fixed), \emph{arm} (the arm tracks the drone, base fixed), and \emph{arm \& body} (the base also steps to re-center the arm).

\begin{figure}[!htbp]
  \centering
  \includegraphics[width=\columnwidth]{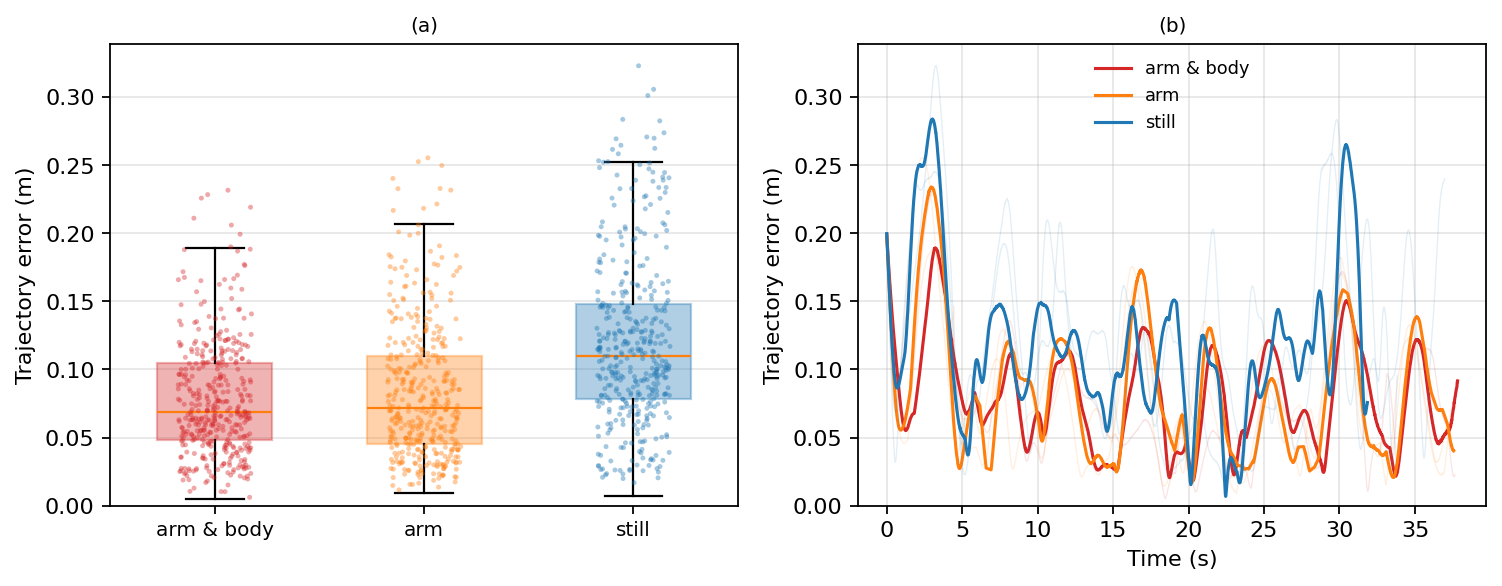}
  \caption{Experiment 3: Planned trajectory tracking error by quadruped motion using 640$\times$480 resolution. (a)~tracking error boxplots per quadruped motion condition; (b)~tracking error over time per quadruped motion condition}
  \label{fig:exp3}
\end{figure}

We report the drone's overall trajectory error per condition (Fig.~\ref{fig:exp3}a) rather than binning by distance, because the conditions sit at different ranges by design: arm tracking holds the camera at its $\approx$0.9\,m stand-off, whereas the still arm sees the drone at $\approx$2.7\,m to make it possible to cover the trajectory space.

Active tracking did not degrade control---it improved it: the median trajectory error was 11.0\,cm with the quadruped still, 7.1\,cm with arm tracking, and 6.9\,cm with arm\,\&\,body tracking. The cause is geometric: tracking holds the camera in the close range where AprilTag localization is most accurate (section \S\ref{sec:exp1}), so the platform's motion improves the pose feeding the drone's controller. The per-flight error stays bounded throughout each run (Fig.~\ref{fig:exp3}b).

\section{DISCUSSION}

The three experiments together locate the system's error budget. The raw AprilTag-vs-MOCAP gap (46\,mm) is mostly fixed calibration error: after alignment the intrinsic accuracy is 12--16\,mm (Experiment~1). Flying on the AprilTag pose costs a median 3--5\,cm of tracking error over MOCAP, yet still completes the trajectory autonomously (Experiment~2). The key finding is geometric: actively tracking the drone does more than keep it in frame---it holds it in the close range where AprilTag is most accurate (closer is better, down to the point where the drone risks leaving the camera's view), so the quadruped's motion \emph{improves} the pose it supplies rather than disturbing it (Experiment~3: 11.0\,cm still vs.\ 6.9\,cm with arm\,\&\,body tracking). The 640$\times$480 choice compounds this, trading a few millimeters of per-detection accuracy for $2.4\times$ the detection rate and far fewer stalls.

\textbf{Latency.} The drone fuses the external pose with its onboard IMU, coasting on the IMU between updates, so a late pose is harmless as long as the gap stays short. The AprilTag pose reaching the controller is a median 72\,ms old (p95 $\approx$105\,ms) versus 2\,ms for MOCAP (Fig.~\ref{fig:exp2}), dominated by image delivery and rectification over the Spot SDK's gRPC interface rather than by tag detection. At the flown speed ($\approx$0.25\,m/s) this is a position lag of roughly 2\,cm, concentrated at the direction reversals---the overshoot in Fig.~\ref{fig:exp2}a. If the pose ever becomes older than a fixed timeout---the external-pose \emph{watchdog} (300\,ms for MOCAP, 500\,ms for AprilTag)---the drone stops trusting it and lands. To tackle these problems, a GPU-based AprilTag detector could be used to speed up detection.

\textbf{Limitations.} The pose is produced in the quadruped's map frame, directly usable by a path planner; we evaluate localization and tracking here rather than planning. The dominant failure mode is the fiducials themselves: detection gaps under rapid yaw, and accuracy loss at distance, oblique angles, or specular glare. This is not architectural---the pipeline only needs $T_{\text{cam}\to \text{drone}}$, so a different markerless detector could replace AprilTag without touching the rest of the system. The drone's working region is a sphere set by the arm's reach and the accurate-detection range. The base following enlarges it, which suits our target regime of slow, close-range flight. The hardest limitation is occlusion: the quadruped's camera cannot see the drone when it passes through the narrow openings. Moreover, we have not run outside the lab (variable lighting, 2.4\,GHz communication constraints, unstructured environments), and the drone's $\approx$7\,min battery caps mission length.

Even with these limits, the system delivers what its setting demands: a planner-ready, metric pose for a drone using only the quadruped's onboard sensing---no motion capture and no fixed infrastructure---in exactly the GNSS-denied, confined environments where that capability is otherwise hard to obtain.

\section{CONCLUSION}

We presented a cooperative robotic system in which a quadruped robot with a 7-DOF arm provides the full 6-DOF pose of a micro-UAV that has no localization sensing of its own, while both robots move. The quadruped calculates its localization, the arm carrying a camera moves to keep the drone in view, the AprilTag pose is calculated relative to the quadruped and then transformed to the map frame, and the drone's onboard EKF combines that pose with its IMU to follow setpoints. Putting all of the localization on the ground robot lets a 42\,g drone (27\,g drone + 15\,g AprilTag cube-like chassis) with no exteroceptive sensors fly under autonomous control.

We tested the system in three lab experiments. The external pose locates the drone to 12--16\,mm after a constant calibration offset is removed (Experiment~1). When the drone flies on this pose, it follows the commanded path with a median error of 8--11\,cm, about 2--5\,cm worse than the same flight flown on a motion-capture pose (Experiment~2). Active arm tracking keeps the drone in the close range where the markers are most accurate, so the quadruped's motion lowers the trajectory error---from 11.0\,cm with the quadruped standing still to 6.9\,cm with the arm and body both following the drone (Experiment~3).

These results come from a lab with motion-capture ground truth and pre-surveyed tags; we have not yet run the system outside the lab or through a real confined opening. We measured localization and tracking, not planning: the pose is already in the quadruped's map frame, so a mapping-and-planning module could consume it, but we have not built or tested one. The main limitations are the markers, whose detections drop under fast yaw and at long or oblique views, and occlusion, since the camera loses the drone when it passes through a narrow gap. Replacing the markers with a markerless detector would remove the tags, and fusing the drone's IMU with the visual pose in one estimator could let the drone hold its pose for short periods while out of view; both are left for future work.

\section*{ACKNOWLEDGMENT}

This research was supported by the IT University of Copenhagen and Helix Lab in Denmark under the Novo Nordisk Fonden through project grant number 224611. We also thank Rasmus Finderup Thomsen for his help at the Helix Lab Robotics Lab.

\bibliographystyle{splncs04}
\bibliography{references}

\begin{thebibliography}{10}
\providecommand{\url}[1]{\texttt{#1}}
\providecommand{\urlprefix}{URL }
\providecommand{\doi}[1]{https://doi.org/#1}

\bibitem{depetrillo_ugv_uav_2024}
Akhihiero, D., Olawoye, U., Das, S., Gross, J.: Cooperative localization for
  {GNSS}-denied subterranean navigation: A {UAV}--{UGV} team approach.
  NAVIGATION: J. Inst. Nav.  \textbf{71}(4),  navi.677 (2024)

\bibitem{bajcsy_active_2018}
Bajcsy, R., Aloimonos, Y., Tsotsos, J.: Revisiting active perception. Auton.
  Robots  \textbf{42}(2),  177--196 (2018)

\bibitem{bdai_spot_ros2}
{BDAI Institute}: \texttt{spot\_ros2} -- {ROS\,2} driver for {Boston Dynamics}
  {Spot}. \url{https://github.com/bdaiinstitute/spot_ros2} (2023)

\bibitem{boston_dynamics_spot}
{Boston Dynamics}: Spot. \url{https://bostondynamics.com/products/spot/},
  accessed 2026

\bibitem{chaumette_vs_2006}
Chaumette, F., Hutchinson, S.: Visual servo control. part~{I}: Basic
  approaches. IEEE Robot. Autom. Mag.  \textbf{13}(4),  82--90 (2006)

\bibitem{chiun_star_2024}
Chiun, J., Tan, Y., Cao, Y., Tan, J., Sartoretti, G.: {STAR}: Swarm technology
  for aerial robotics research. arXiv:2406.16671 (2024)

\bibitem{geneva_openvins_2020}
Geneva, P., Eckenhoff, K., Lee, W., Yang, Y., Huang, G.: {OpenVINS}: A research
  platform for visual-inertial estimation. In: Proc. IEEE {ICRA}. pp.
  4666--4672 (2020)

\bibitem{gross_aerial_2019}
Gross, J., et~al.: Field-testing of a {UAV}--{UGV} team for {GNSS}-denied
  navigation in subterranean environments. In: Proc. 32nd {ION} {GNSS+}. pp.
  2112--2124 (2019)

\bibitem{honig_crazyswarm2}
H{\"o}nig, W., et~al.: {Crazyswarm2}: A {ROS\,2}-based stack for {Bitcraze}
  {Crazyflie} multirotor robots. \url{https://github.com/IMRCLab/crazyswarm2}
  (2023)

\bibitem{horn_1987}
Horn, B.: Closed-form solution of absolute orientation using unit quaternions.
  J. Opt. Soc. Am. A  \textbf{4}(4),  629--642 (1987)

\bibitem{jarraya_gnss_denied_2025}
Jarraya, I., et~al.: {GNSS}-denied unmanned aerial vehicle navigation:
  Analyzing computational complexity, sensor fusion, and localization
  methodologies. Satellite Navigation  \textbf{6}(1), ~9 (2025)

\bibitem{kalaitzakis_fiducial_2021}
Kalaitzakis, M., et~al.: Fiducial markers for pose estimation: Overview,
  applications and experimental comparison of the {ARTag}, {AprilTag}, {ArUco}
  and {STag} markers. J. Intell. Robot. Syst.  \textbf{101}(4), ~71 (2021)

\bibitem{macenski_ros2_2022}
Macenski, S., Foote, T., Gerkey, B., Lalancette, C., Woodall, W.: Robot
  operating system 2: Design, architecture, and uses in the wild. Sci. Robot.
  \textbf{7}(66),  eabm6074 (2022)

\bibitem{markley_2007}
Markley, F., Cheng, Y., Crassidis, J., Oshman, Y.: Averaging quaternions. J.
  Guid. Control Dyn.  \textbf{30}(4),  1193--1197 (2007)

\bibitem{olson_apriltag_2011}
Olson, E.: {AprilTag}: A robust and flexible visual fiducial system. In: Proc.
  IEEE {ICRA}. pp. 3400--3407 (2011)

\bibitem{palossi_nanodrone_2019}
Palossi, D., Loquercio, A., Conti, F., Flamand, E., Scaramuzza, D., Benini, L.:
  A {64-mW} {DNN}-based visual navigation engine for autonomous nano-drones.
  IEEE Internet Things J.  \textbf{6}(5),  8357--8371 (2019)

\bibitem{qin_vinsmono_2018}
Qin, T., Li, P., Shen, S.: {VINS-Mono}: A robust and versatile monocular
  visual-inertial state estimator. IEEE Trans. Robot.  \textbf{34}(4),
  1004--1020 (2018)

\bibitem{queralta_uwb_2020}
Shule, W., et~al.: {UWB}-based localization for multi-{UAV} systems and
  collaborative heterogeneous multi-robot systems: A survey. Procedia Comput.
  Sci.  \textbf{175},  357--364 (2020)

\bibitem{xu_fastlio2_2022}
Xu, W., Cai, Y., He, D., Lin, J., Zhang, F.: {FAST-LIO2}: Fast direct
  {LiDAR}-inertial odometry. IEEE Trans. Robot.  \textbf{38}(4),  2053--2073
  (2022)

\end{thebibliography}

\end{document}